\documentclass[sigconf]{acmart}
\usepackage{makecell}
\usepackage{multirow}
\usepackage{graphicx}
\usepackage{subfigure}
\usepackage[normalem]{ulem}
\usepackage{url}
\usepackage{enumerate}
\usepackage{booktabs}
\usepackage{multirow}
\usepackage{xspace}

\usepackage{amsmath}
\usepackage{color}
\usepackage{threeparttable}
\usepackage{ulem}
\usepackage{amssymb}

\usepackage{float}
\usepackage[ruled,linesnumbered]{algorithm2e}
\usepackage{algorithm2e, setspace, algpseudocode}

\def\bf#1{\mathbf{#1}}
\def\cal#1{\mathcal{#1}}

\newcommand{\eg}{\textit{e.g.,}\xspace}
\newcommand{\ie}{\textit{i.e.,}\xspace}

\newtheorem{definition}{Definition}

\providecommand{\tabularnewline}{\\}

\AtBeginDocument{%
  \providecommand\BibTeX{{%
    \normalfont B\kern-0.5em{\scshape i\kern-0.25em b}\kern-0.8em\TeX}}}

\copyrightyear{2022}
\acmYear{2022}
\setcopyright{rightsretained}
\acmConference[CIKM '22]{Proceedings of the 31st ACM International Conference on Information and Knowledge Management}{October 17--21, 2022}{Atlanta, GA, USA}
\acmBooktitle{Proceedings of the 31st ACM International Conference on Information and Knowledge Management (CIKM '22), October 17--21, 2022, Atlanta, GA, USA}
\acmDOI{10.1145/3511808.3557702}
\acmISBN{978-1-4503-9236-5/22/10}

\usepackage{etoolbox}
\makeatletter
\patchcmd{\maketitle}{\@copyrightpermission}{
$^\dagger$ Zezhi Shao is also with the University of Chinese Academy of Sciences.\\
$^\ddagger$ Fei Wang is the corresponding author.\\
   \begin{minipage}{0.3\columnwidth}
     \href{https://creativecommons.org/licenses/by/4.0/}{\includegraphics[width=0.90\textwidth]{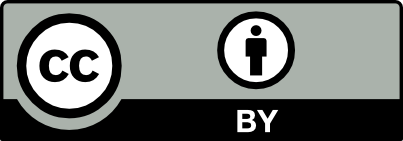}}
   \end{minipage}\hfill
   \begin{minipage}{0.7\columnwidth}
     \href{https://creativecommons.org/licenses/by/4.0/}{
     This work is licensed under a Creative Commons Attribution International 4.0 License.}
   \end{minipage}
  
   \vspace{5pt}
}{}{}

\makeatother

\begin{document}

\title{Spatial-Temporal Identity: A Simple yet Effective Baseline for Multivariate Time Series Forecasting}




\author{Zezhi Shao$^\dagger$\\ Zhao Zhang \\ Fei Wang$^\ddagger$}
\affiliation{
\institution{Institute of Computing Technology, \\Chinese Academy of Sciences}
\country{}
}
\email{{shaozezhi19b, zhangzhao2021, wangfei}@ict.ac.cn}



\author{Wei Wei}
\affiliation{
\institution{Huazhong University of Science and Technology}
\country{}
}
\email{wangfei@ict.ac.cn}

\author{Yongjun Xu}
\affiliation{
\institution{Institute of Computing Technology, \\Chinese Academy of Sciences}
\country{}
}
\email{xyj@ict.ac.cn}

\renewcommand{\shortauthors}{Zezhi Shao et al.}
\renewcommand{\authors}{Zezhi Shao, Zhao Zhang, Fei Wang, Wei Wei, Yongjun Xu}


\begin{abstract}

Multivariate Time Series~(MTS) forecasting plays a vital role in a wide range of applications.
Recently, Spatial-Temporal Graph Neural Networks~(STGNNs) have become increasingly popular MTS forecasting methods due to their state-of-the-art performance.
However, recent works are becoming more sophisticated with limited performance improvements.
This phenomenon motivates us to explore the critical factors of MTS forecasting and design a model that is as powerful as STGNNs, but more concise and efficient.
In this paper, we identify the indistinguishability of samples in both spatial and temporal dimensions as a key bottleneck, and propose a simple yet effective baseline for MTS forecasting by attaching \underline{S}patial and \underline{T}emporal \underline{ID}entity information~(STID), which achieves the best performance and efficiency simultaneously based on simple Multi-Layer Perceptrons~(MLPs).
These results suggest that we can design efficient and effective models as long as they solve the indistinguishability of samples, without being limited to STGNNs.

\end{abstract}



\maketitle

\section{Introduction}

\begin{figure}[t]
  \centering
  \setlength{\abovecaptionskip}{0.2cm}
  \setlength{\belowcaptionskip}{-0.4cm}
  \begin{minipage}[h]{\linewidth}
  \hspace{-0.1cm}\includegraphics[width=1\linewidth]{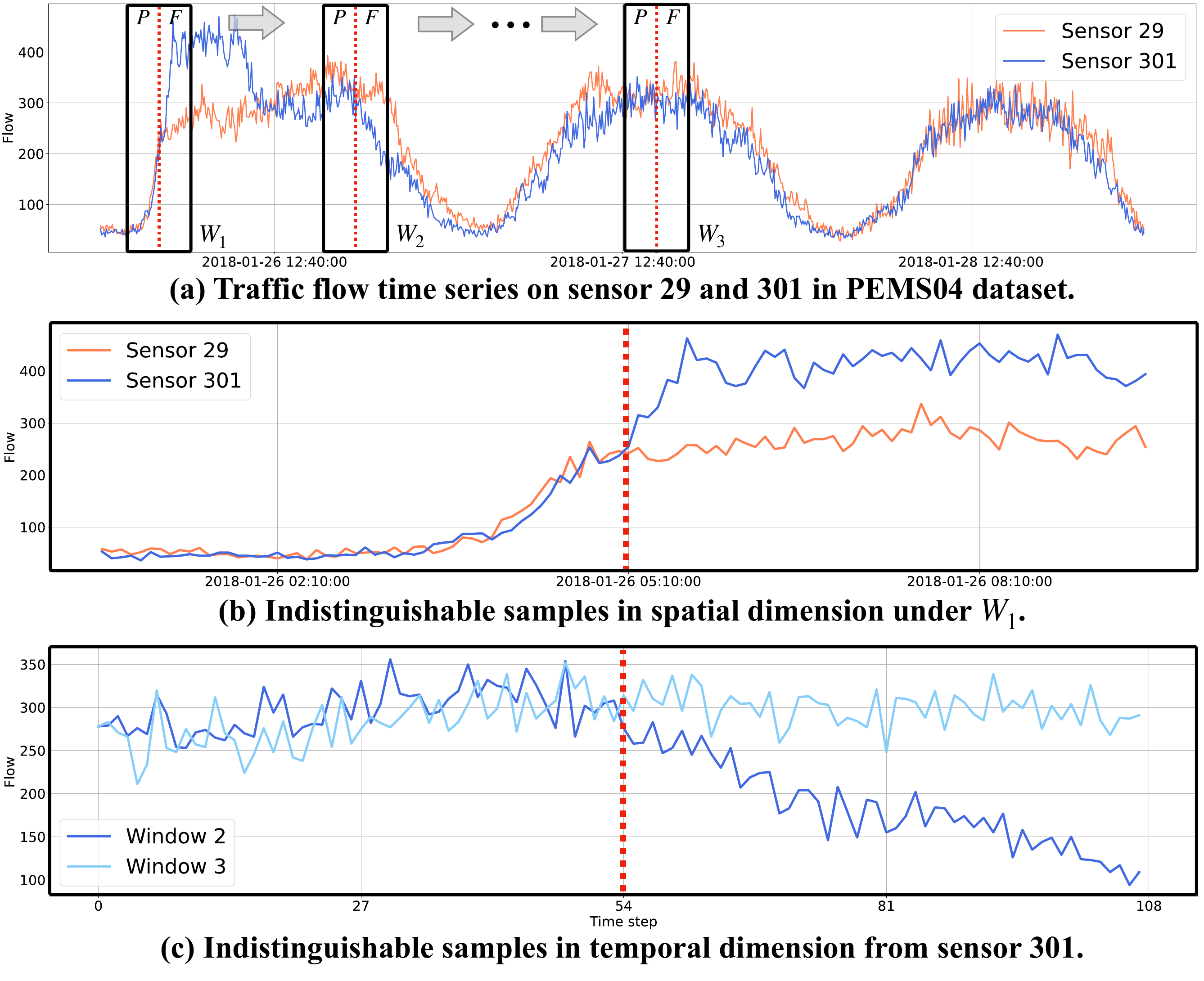}
  \caption{Examples of traffic flow MTS data and the indistinguishable samples in the spatial and temporal dimension.}
  \label{Intro}
  \end{minipage}
\end{figure}

Multivariate time series~(MTS) data is a typical spatial-temporal data, which contains multiple interrelated time series. 
Accurate and efficient MTS forecasting plays a vital role in many applications, from transportation and energy to economics~\cite{2017DCRNN, 2021STNorm, 2021Stock}, and has remained an enduring research topic in both academia and industry.

Previous studies on MTS forecasting usually fall into two categories, \ie statistical methods and deep learning-based methods.
The former assume that there exist linear correlations among variables~(\ie time series).
%
Regarding the latter, early works~\cite{2018LSTNet} utilize Convolution Neural Networks~(CNN) to capture the correlations among variables, yet ignore their non-Euclidean pairwise dependencies.
%
Recently, Spatial-Temporal Graph Neural Networks~(STGNNs)~\cite{GWNet, D2STGNN, STEP} have attracted increasing attention for their state-of-the-art performance.
STGNNs combine graph convolutional networks~(GCN~\cite{2017GCN}) and sequential models~\cite{2014GRU, 2016TCN}. 
The former deals with non-Euclidean dependencies among variables, and the latter captures temporal patterns.
Many researchers have made persistent efforts to design powerful graph convolutions~\cite{2018STGCN, 2020GMAN, 2020StemGNN}, or to reduce reliance on the pre-defined graph structure~\cite{2020MTGNN, 2021REST, 2021GTS}.
Despite significant progress, recent STGNN-based methods are becoming sophisticated with limited improvements, which motivates us to think: \textit{can we refine the critical factors of MTS forecasting, and design a model that is as powerful as STGNNs but more concise and efficient?}
To answer the above question, in this paper, we first identify the indistinguishability of samples in both spatial and temporal dimensions as a key bottleneck. 
Subsequently, we design a simple yet effective baseline model to alleviate this bottleneck.

To intuitively illustrate our observation, we take the MTS data in Figure \ref{Intro} as an example, where each time series is derived from a traffic flow sensor.
First of all, as shown in Figure \ref{Intro}(a), the samples are generated by a sliding window with a size of $P+F$, where $P$ and $F$ denote the length of historical data and future data, respectively. 
For example, $W_1$, $W_2$, and $W_3$ are three windows at different time.
Furthermore, considering that different variables and periods have different patterns, we can expect to generate many samples with similar historical data but different future data.
For example, Figure \ref{Intro}(b) shows samples from different variables~(\ie sensors 29 and 301) under window $W_1$, where the historical data~(left) are very similar and the future data~(right) are different.
Similarly, samples from sensor 301 under different periods~(\ie windows $W_2$ and $W_3$) are shown in Figure \ref{Intro}(c).
Simple regression models~(\eg MLPs) cannot predict their different future data based on their similar historical data, that is, they can not distinguish these samples.
Therefore, we refer to the characteristics behind the two kinds of sample pairs in Figures \ref{Intro}(b) and \ref{Intro}(c) as the indistinguishability of samples in the spatial and temporal dimensions.
In addition, a very recent work~\cite{2021STNorm} also reveals that the critical factor for the success of STGNNs is that GCN relieves spatial indistinguishability.

To alleviate the above bottleneck, we design a simple yet effective baseline model for MTS forecasting, named STID, based on an intuitive idea of attaching spatial-temporal identity information.
As shown in Figure \ref{STID}, STID utilizes a spatial embedding matrix $\mathbf{E}\in\mathbb{R}^{N\times D}$, and two temporal embedding matrices $\mathbf{T}^{\text{TiD}}\in\mathbb{R}^{N_d\times D}$ and $\mathbf{T}^{\text{DiW}}\in\mathbb{R}^{N_w\times D}$, to indicate the spatial and temporal identities.
$N$ is the number of variables~(\ie time series), $N_d$ is the number of time slots in a day~(determined by the sampling frequency), $N_w=7$ is the number of days in a week, and $D$ is the hidden dimension.
Subsequently, STID encodes information based on simple MLP layers and makes predictions through a regression layer.
%
STID has a more concise architecture compared with the STGNN-based methods, and extensive experiments have shown that STID is more powerful than STGNN-based methods and has significant efficiency advantages.
These results suggest that we can design more efficient and effective models by solving the indistinguishability of samples, without being limited to STGNNs.

\section{Preliminaries}
\begin{definition}
\textbf{Multivariate Time Series Forecasting.}
Multivariate time series can be denoted as a tensor $\mathbf{X}\in\mathbb{R}^{T\times N}$, where $T$ is the number of time slots and $N$ is the number of variables.
Given historical signals $\mathbf{X}\in\mathbb{R}^{P\times N}$ from the past $P$ time slots, multivariate time series forecasting aims to predict the values $\mathbf{Y}\in\mathbb{R}^{F\times N}$ of the $F$ nearest future time slots.
We additionally denote the sample from time series $i$ at time step $t$ as $\mathbf{X}^i_{t-P:t}\in\mathbb{R}^{P}$ and $\mathbf{Y}^i_{t:t+F}\in\mathbb{R}^{F}$.
\end{definition}

\begin{definition}
\textbf{Spatial and Temporal Identities.}
Assuming $N$ time series and $N_d$ time slots in a day and $N_w=7$ days in a week, the spatial and temporal identities are preserved in three embedding matrices, \ie $\mathbf{E}\in\mathbb{R}^{N\times D}$, $\mathbf{T}^{\text{TiD}}\in\mathbb{R}^{N_d\times D}$, and $\mathbf{T}^{\text{DiW}}\in\mathbb{R}^{N_w\times D}$, which are trainable parameters, and $D$ is the hidden dimension.
\end{definition}

\section{Model Architecture}
\begin{figure*}[ht]
  \centering
  \setlength{\abovecaptionskip}{0.2cm}
  \setlength{\belowcaptionskip}{-0.1cm}
  \includegraphics[width=1\linewidth]{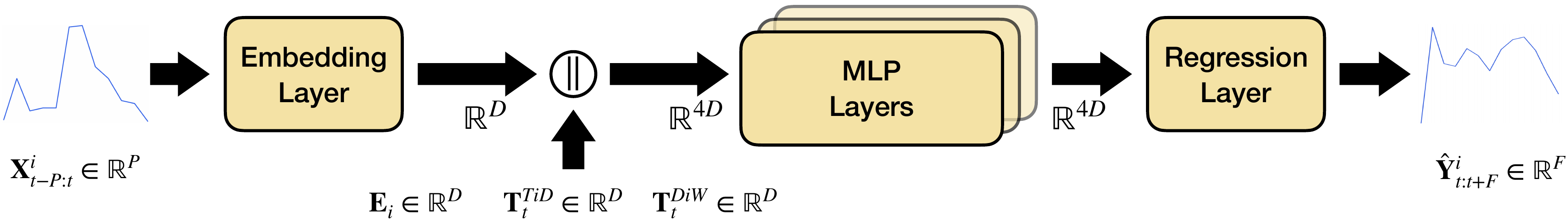}
  \caption{The overview of the proposed STID.}
  \label{STID}
\end{figure*}
As shown in Figure \ref{STID}, STID consists of an embedding layer, multiple MLP layers, and a regression layer. 
For simplicity, we denote $\text{FC}(\cdot)$ as a fully connected layer.
The embedding layer transforms raw historical time series $\mathbf{X}_{t-P:t}^i\in\mathbb{R}^{P}$ into latent space $\mathbf{H}_t^i\in\mathbb{R}^{D}$ by:
\begin{equation}
    \mathbf{H}^i_t = \text{FC}_{\text{embedding}}(\mathbf{X}_{t-P:t}^i),
\end{equation}
where $D$ is the hidden dimension.
Then, STID attaches spatial and temporal identities $\mathbf{E}_i$, $\mathbf{T}^{\text{TiD}}$, and $\mathbf{T}^{\text{DiW}}$ by:
\begin{equation}
    \mathbf{Z}^i_t = \mathbf{H}^i_t \parallel \mathbf{E}_i \parallel \mathbf{T}^{\text{TiD}}_t \parallel \mathbf{T}^{\text{DiW}}_t,
\end{equation}
where $\mathbf{Z}^i_t\in\mathbb{R}^{4D}$ denotes the hidden representation with spatial and temporal identities.
Kindly note that the spatial and temporal identities are randomly initialized trainable parameters, and the temporal identities are shared among time slots for the same time in a day and the same day in a week.
Subsequently, we utilize $L$ layers of MLP with a residual connection to encode information.
The $l$-th MLP layer can be denoted as:
\begin{equation}
    (\mathbf{Z}^i_t)^{l+1} = \text{FC}_2^l(\sigma(\text{FC}_1^l((\mathbf{Z}^i_t)^{l}))) + (\mathbf{Z}^i_t)^{l}.
\end{equation}
Finally, the regression layer makes predictions based on $(\mathbf{Z}^i_t)^{L}$:
\begin{equation}
    \hat{\mathbf{Y}}^i_{t:t+F} = \text{FC}_{\text{regression}}((\mathbf{Z}^i_t)^{L}),
\end{equation}
where $(\mathbf{Z}^i_t)^{L}\in\mathbb{R}^{4D}$, and $\hat{\mathbf{Y}}^i_{t:t+F}\in\mathbb{R}^{F}$ is the prediction.
We use Mean Absolute Error~(MAE) as our loss function:
\begin{equation}
    \cal{L}(\hat{\bf{Y}}, \bf{Y})=\frac{1}{NF}
    \sum_{i=1}^{N}\sum_{j=1}^{F}|\hat{\bf{Y}}^i_j - \bf{Y}^i_{j}|.
    \label{loss}
\end{equation}
We optimize the parameters of all spatial and temporal identities and fully connect layers by minimizing $\mathcal{L}$ via backpropagation and gradient descent.
We choose Adam~\cite{Adam} as our optimizer.

\section{Experiments}
\subsection{Experimental Setup}
\noindent\textbf{Datasets.}
Following previous works~\cite{2020MTGNN, GWNet, 2021STNorm}, we conduct experiments on five commonly used multivariate time series datasets: PEMS04, PEMS07, PEMS08, PEMS-BAY, and Electricity.
The statistical information is summarized in Table \ref{tab:datasets}.
%
%
%
It is notable that PEMS04, PEMS07, PEMS08, and PEMS-BAY datasets come with a pre-defined graph to indicate the dependencies among time series.
Due to space limitations, we do not introduce each dataset in detail.

\begin{table}
\renewcommand\arraystretch{0.9}
\setlength{\abovecaptionskip}{0.cm}
\setlength{\belowcaptionskip}{-0.0cm}
\caption{Statistics of datasets.}
\label{tab:datasets}
\scalebox{0.93}{
  \begin{tabular}{c|c|c|c|c}    
    \toprule
    \textbf{Dataset} &\textbf{Length} & \textbf{\# Variants} & \textbf{Sample Rate} & \textbf{Time Span}\\
    \midrule
    {PEMS04}   & 16992 & 307 & 5mins & 2 months\\
    {PEMS07}   & 28224 & 883 & 5mins & 3 months\\
    {PEMS08}   & 17856 & 170 & 5mins & 2 months\\
    {PEMS-BAY} & 52116 & 325 & 5mins & 6 months\\
    {Electricity}   & 2208 & 336 & 60mins & 3 months\\
    \bottomrule
  \end{tabular}
}
\end{table}
\setlength{\textfloatsep}{5pt}
\noindent\textbf{Baselines.} 
We select a wealth of baselines that have official public code, including the traditional methods~(VAR~\cite{VAR}, HI~\cite{2021HI}) and the typical deep learning methods~(LSTM~\cite{1997LSTM}, DCRNN~\cite{2017DCRNN}, STGCN~\cite{2018STGCN}, Graph WaveNet~\cite{GWNet}, AGCRN~\cite{2020AdaptiveGCRN}, StemGNN~\cite{2020StemGNN}), as well as  the very recent works~(GMAN~\cite{2020GMAN}, MTGNN~\cite{2020MTGNN}, ST-Norm~\cite{2021STNorm}).
Due to space limitations, we do not introduce each method in detail.

\noindent\textbf{Metrics.}
We evaluate the performances of all baselines by three commonly used metrics in multivariate time series forecasting, including Mean Absolute Error (MAE), Root Mean Squared Error (RMSE), and Mean Absolute Percentage Error (MAPE).

\noindent\textbf{Implementation.}
The proposed model is implemented with Pytorch 1.9.1 on an NVIDIA RTX 2080Ti GPU.
The hidden dimension $D$ is set to $32$.
The number of MLP layers $L$ is set to $3$.
For PEMS04, PEMS07, PEMS08, and PEMS-BAY datasets, we set the length of historical data $P$ to 12. For the Electricity dataset, we set $P=168$.
For all datasets, we set the length of future data $F$ to 12.
The learning rate is set to 0.001.
The code is available at \url{https://github.com/zezhishao/STID}.

\subsection{Performance Study}
For a fair comparison, we follow the dataset division in previous works.
The ratio of training, validation, and test sets for the PEMS-BAY dataset is 7 : 1 : 2, while the ratio for other datasets is 6 : 2 : 2.
We aim to predict the future time series with a length of 12, \ie $F=12$, on all datasets.
The results are shown in Table \ref{tab:main}.
We compared the performance of these methods on the 3rd, 6th, and 12th time slots as well as the performance of the average 12 time slots, which are shown in the \textbf{@3}, \textbf{@6}, \textbf{@12}, and \textbf{avg} columns, respectively.
The best results are highlighted in bold, and the second-best results are underlined.
In addition, DCRNN, STGCN, Graph WaveNet~(\ie GWNet), and GMAN rely on a pre-defined graph.
Therefore, since there is no graph structure, the results of these methods in the Electricity dataset are not available.
As shown in the table, STID consistently achieves the best performance in almost all horizons in all datasets and does not require a pre-defined graph.
These remarkable results demonstrate the effectiveness of STID.

\subsection{Efficiency Study}
In this part, we compare the efficiency of STID with other learning methods based on all datasets.
For a more intuitive and effective comparison, we compare the average training time required for each epoch of these models.
All models are trained on Intel(R) Xeon(R) Gold 5217 CPU @ 3.00GHz, 128G RAM computing server, equipped with NVIDIA RTX 2080Ti graphics cards.

The results are shown in Table \ref{tab:efficiency}.
The computational complexity of previous STGNN-based models usually increases linearly or quadratically with the length of the input time series and the number of variables.
Compared with other datasets, the PEMS07 dataset has more variables~($N=883$), and the Electricity dataset has longer historical data~($P=168$).
Therefore, previous works spend more time on the PEMS07 and Electricity datasets.
Thanks to the concise architecture without GCN and sequential models~(\eg RNN), STID achieves consistent best efficiency on all datasets.

\begin{table}
\renewcommand\arraystretch{1}
    \centering
    \setlength{\abovecaptionskip}{0.cm}
    \setlength{\belowcaptionskip}{-0.0cm}
    \caption{Efficiency study.}
    \label{tab:efficiency}

\scalebox{0.835}{
\begin{tabular}{c|c|c|c|c|c}
\toprule 
\midrule 
\textbf{Dataset} & \textbf{PEMS04} & \textbf{PEMS07} & \textbf{PEMS08} & \textbf{PEMS-BAY} & \textbf{Electricity}\tabularnewline
\midrule 
\midrule 
\textbf{Methods} & \multicolumn{5}{c}{\textbf{Seconds/epoch}}\tabularnewline
\midrule 
VAR & 14.73 & 189.37 & 7.65 & 57.11 & 29.62\tabularnewline
\midrule 
LSTM & 7.78 & 25.73 & 4.56 & 28.34 & 22.26\tabularnewline
\midrule 
DCRNN & 95.12 & 510.53 & 57.17 & 351.35 & N/A\tabularnewline
\midrule 
STGCN & 41.16 & 198.13 & 25.31 & 155.68 & N/A\tabularnewline
\midrule 
GWNet & 27.88 & 170.61 & 29.72 & 111.95 & N/A\tabularnewline
\midrule 
AGCRN & 28.49 & 189.50 & 19.29 & 102.09 & 144.17\tabularnewline
\midrule 
StemGNN & 16.29 & 136.31 & 9.41 & 63.791 & 56.24\tabularnewline
\midrule 
GMAN & 107.31 & 827.77 & 71.04 & 410.67 & N/A\tabularnewline
\midrule 
MTGNN & 25.11 & 107.03 & 79.45 & 90.18 & 28.59\tabularnewline
\midrule 
STNorm & 18.20 & 74.12 & 32.45 & 64.36 & 155.86\tabularnewline
\midrule 
\midrule 
STID & \textbf{5.24} & \textbf{14.32} & \textbf{4.46} & \textbf{15.76} & \textbf{2.31}\tabularnewline
\midrule 
\bottomrule
\end{tabular}
}
\end{table}

\subsection{Ablation Study}
\begin{figure}[h]
    \setlength{\abovecaptionskip}{0.2cm}
    \setlength{\belowcaptionskip}{-0.4cm}
  \centering
  \includegraphics[width=1\linewidth]{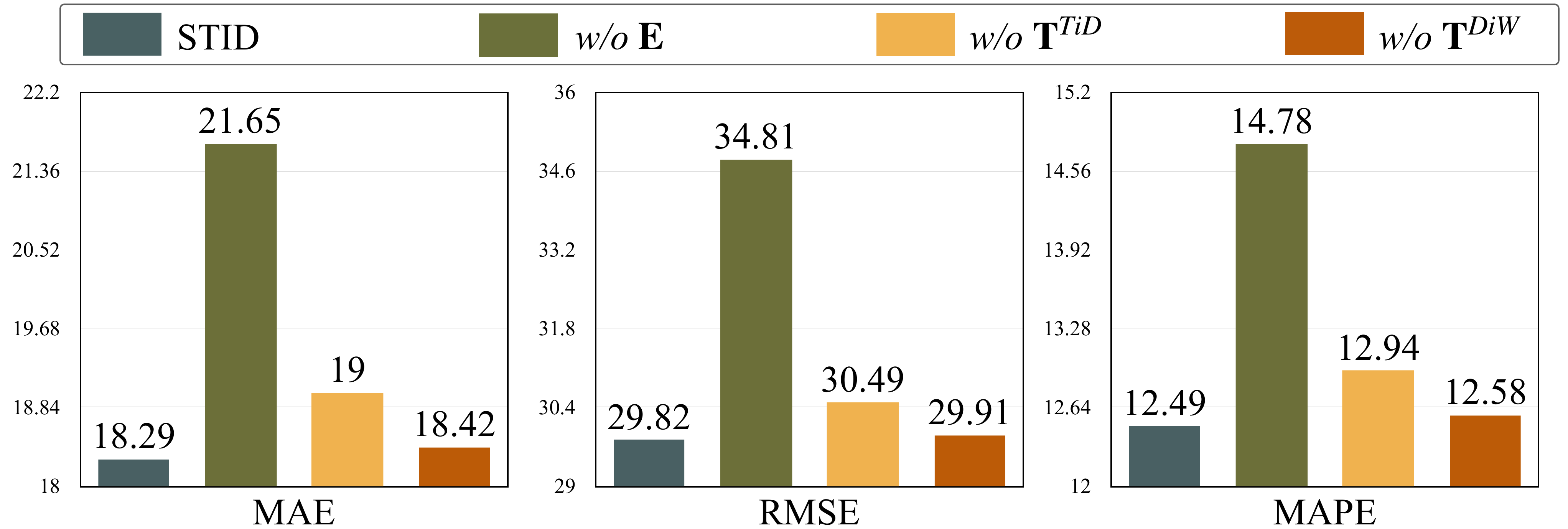}
  \caption{Ablation study on PEMS04 dataset.}
  \label{ablation}
\end{figure}

In this part, we conduct ablation studies to verify the effectiveness of spatial-temporal identity.
We set three variants of our STID.
\textit{STID w/o} $\mathbf{E}$ removes the spatial identity.
\textit{STID w/o} $\mathbf{T}^{\text{TiD}}$ removes the $\mathbf{T}^{\text{TiD}}$ temporal identity, while \textit{STID w/o} $\mathbf{T}^{\text{DiW}}$ removes the $\mathbf{T}^{\text{DiW}}$ temporal identity.
We conduct experiments on the PEMS04 dataset and report the \textbf{avg} column on all metrics.
The results are shown in Figure \ref{ablation}.
In summary, all these identities are beneficial. 
The most important one is spatial identity, which means that spatial indistinguishability acts as a major bottleneck of MTS forecasting.
Furthermore, the temporal identities $\mathbf{T}^{\text{TiD}}$ and $\mathbf{T}^{\text{DiW}}$ are also important since data in the real world often contain daily and weekly periodicity.

\begin{table*}
\renewcommand\arraystretch{1}
    \centering
    \setlength{\abovecaptionskip}{0.cm}
    \setlength{\belowcaptionskip}{-0.0cm}
    \caption{Multivariate time series forecasting on the PEMS04, PEMS07, PEMS08, PEMS-BAY, and Electricity datasets. }
    \label{tab:main}

\scalebox{0.64}{
    \begin{tabular}{cc|cccc|cccc|cccc|cccc|cccc}
    \toprule 
    \midrule
    \multicolumn{2}{c|}{\textbf{Dataset}} & \multicolumn{4}{c|}{\textbf{PEMS04}} & \multicolumn{4}{c|}{\textbf{PEMS07}} & \multicolumn{4}{c|}{\textbf{PEMS08}}& \multicolumn{4}{c|}{\textbf{PEMS-BAY}} & \multicolumn{4}{c}{\textbf{Electricity}}\tabularnewline
    \midrule 
    \midrule 
    \textbf{Method} & \textbf{Metric} & \textbf{@3} & \textbf{@6} & \textbf{@12} & \textbf{Avg.} & \textbf{@3} & \textbf{@6} & \textbf{@12} & \textbf{Avg.} & \textbf{@3} & \textbf{@6} & \textbf{@12} & \textbf{Avg.} & \textbf{@3} & \textbf{@6} & \textbf{@12} & \textbf{Avg.} & \textbf{@3} & \textbf{@6} & \textbf{@12} & \textbf{Avg.}\tabularnewline
    \midrule 
\multirow{3}{*}{HI} & MAE & 42.33 & 42.35 & 42.37 & 42.36 & 49.02 & 49.03 & 49.06 & 49.04 & 34.55 & 34.57 & 34.59 & 34.57 & 3.06 & 3.06 & 3.05 & 3.06 & 92.44 & 92.58 & 92.79 & 92.58\tabularnewline
 & RMSE & 61.64 & 61.66 & 61.67 & 61.66 & 71.16 & 71.18 & 71.20 & 71.18 & 50.41 & 50.43 & 50.44 & 50.43 & 7.05 & 7.05 & 7.03 & 7.04 & 167.00 & 167.05 & 167.21 & 167.07\tabularnewline
 & MAPE & 29.90\% & 29.92\% & 29.96\% & 29.92\% & 22.73\% & 22.75\% & 22.79\% & 22.75\% & 21.60\% & 21.63\% & 21.68\% & 21.63\% & 6.85\% & 6.84\% & 6.83\% & 6.84\% & 70.16 & 70.46 & 70.91 & 70.43\tabularnewline
\midrule 
\multirow{3}{*}{VAR} & MAE & 21.94 & 23.72 & 26.76 & 23.51 & 32.02 & 35.18 & 38.37 & 37.06 & 19.52 & 22.25 & 26.17 & 22.07 & 1.74 & 2.32 & 2.93 & 2.21 & 27.69 & 28.19 & 29.34 & 28.29\tabularnewline
 & RMSE & 34.30 & 36.58 & 40.28 & 36.39 & 48.83 & 52.91 & 56.82 & 55.73 & 29.73 & 33.30 & 38.97 & 31.02 & 3.16 & 4.25 & 5.44 & 4.12 & 56.06 & 57.55 & 60.45 & 57.78\tabularnewline
 & MAPE & 16.42\% & 18.02\% & 20.94\% & 17.85\% & 18.30\% & 20.54\% & 22.04\% & 19.93\% & 12.54\% & 14.23\% & 17.32\% & 14.04\% & 3.60\% & 5.00\% & 6.50\% & 5.01\% & 75.53\% & 79.94\% & 86.62\% & 80.23\%\tabularnewline
\midrule 
\multirow{3}{*}{LSTM} & MAE & 21.37 & 23.72 & 26.76 & 23.81 & 20.42 & 23.18 & 28.73 & 23.54 & 17.38 & 21.22 & 30.69 & 21.31 & 2.05 & 2.20 & 2.37 & 2.18 & 18.57 & 20.68 & 23.79 & 20.42\tabularnewline
 & RMSE & 33.31 & 36.58 & 40.28 & 36.62 & 33.21 & 37.54 & 45.63 & 38.20 & 26.27 & 31.97\% & 43.96 & 32.10 & 4.19 & 4.55 & 4.96 & 4.47 & 48.86 & 48.96 & 56.44 & 49.03\tabularnewline
 & MAPE & 15.21\% & 18.02\% & 20.94\% & 18.12\% & 8.79\% & 9.80\% & 12.23\% & 9.96\% & 12.63\% & 17.32\% & 25.72\% & 17.47\% & 4.80\% & 5.20\% & 5.70\% & 5.04\% & 32.88\% & \textbf{37.21\%} & \textbf{39.42\%} & \uline{35.58\%}\tabularnewline
\midrule 
\multirow{3}{*}{DCRNN} & MAE & 18.53 & 19.65 & 21.67 & 19.71 & 19.45 & 21.18 & 24.14 & 21.20 & 14.16 & 15.24 & 17.70 & 15.26 & \uline{1.31} & 1.67 & 1.99 & 1.62 & \multirow{1}{*}{} & \multicolumn{2}{c}{N/A} & \tabularnewline
 & RMSE & 29.61 & 31.37 & 34.19 & 31.43 & 31.39 & 34.42 & 38.84 & 34.43 & 22.20 & 24.26 & 27.14 & 24.28 & \textbf{2.80} & 3.81 & 4.66 & 3.74 & \multirow{1}{*}{} & \multicolumn{2}{c}{N/A} & \tabularnewline
 & MAPE & 12.71\% & 13.45\% & 15.03\% & 13.54\% & 8.29\% & 9.01\% & 10.42\% & 9.06\% & 9.31\%\% & 9.90\% & 11.13\% & 9.96\% & \uline{2.73\%} & 3.75\% & 4.73\% & 3.61\% & \multirow{1}{*}{} & \multicolumn{2}{c}{N/A} & \tabularnewline
\midrule 
\multirow{3}{*}{STGCN} & MAE & 18.74 & 19.64 & 21.12 & 19.63 & 20.33 & 21.66 & 24.16 & 21.71 & 14.95 & 15.92 & 17.65 & 15.98 & 1.35 & 1.69 & 2.01 & 1.63 & \multirow{1}{*}{} & \multicolumn{2}{c}{N/A} & \tabularnewline
 & RMSE & 29.84 & 31.34 & 33.53 & 31.32 & 32.73 & 35.35 & 39.48 & 35.41 & 23.48 & 25.36 & 28.03 & 25.37 & 2.88\% & 3.83 & 4.56 & 3.73 & \multirow{1}{*}{} & \multicolumn{2}{c}{N/A} & \tabularnewline
 & MAPE & 14.42\% & 13.27\% & 14.22\% & 13.32\% & 8.68\% & 9.16\% & 10.26\% & 9.25\% & 9.87\% & 10.42\% & 11.34\% & 10.43\% & 2.88\% & 3.85\% & 4.74\% & 3.69\% & \multirow{1}{*}{} & \multicolumn{2}{c}{N/A} & \tabularnewline
\midrule 
\multirow{3}{*}{GWNet} & MAE & \uline{18.00} & 18.96 & 20.53 & 18.97 & \uline{18.69} & \uline{20.26} & 22.79 & \uline{20.25} & \uline{13.72} & 14.67 & 16.15 & \uline{14.67} & \textbf{1.30} & \uline{1.63} & 1.95 & \uline{1.58} & \multirow{1}{*}{} & \multicolumn{2}{c}{N/A} & \tabularnewline
 & RMSE & \uline{28.83} & \uline{30.33} & 32.54 & \uline{30.32} & \uline{30.69} & 33.37 & 37.11 & 33.32 & \uline{21.71} & \textbf{23.50} & \uline{25.95} & \textbf{23.49} & \textbf{2.78} & \uline{3.73} & 4.52 & \uline{3.65} & \multirow{1}{*}{} & \multicolumn{2}{c}{N/A} & \tabularnewline
 & MAPE & 13.64\% & 14.23\% & 15.41\% & 14.26\% & \uline{8.02\%} & \uline{8.56\%} & 9.73\% & \uline{8.63\%} & \uline{8.80\%} & \uline{9.49\%} & 10.74\% & \uline{9.52\%} & \textbf{2.71\%} & \textbf{3.66\%} & 4.63\% & \uline{3.52\%} & \multirow{1}{*}{} & \multicolumn{2}{c}{N/A} & \tabularnewline
\midrule 
\multirow{3}{*}{AGCRN} & MAE & 18.52 & 19.45 & 20.64 & 19.36 & 19.31 & 20.70 & 22.74 & 20.64 & 14.51 & 15.66 & 17.49 & 15.65 & 1.37 & 1.70 & 1.99 & 1.63 & 22.88 & 24.47 & 27.24 & 23.88\tabularnewline
 & RMSE & 29.79 & 31.45 & 33.31 & 31.28 & 31.68 & 34.52 & 37.94 & 34.39 & 22.87 & 25.00 & 27.93 & 24.99 & 2.93 & 3.89 & 4.64 & 3.78 & 49.98 & 54.17 & 59.76 & 53.02\tabularnewline
 & MAPE & 12.31\% & 12.82\% & 13.74\% & 12.81\% & 8.18\% & 8.66\% & 9.71\% & 8.74\% & 9.34\% & 10.34\% & 11.72\% & 10.17\% & 2.95\% & 3.88\% & 4.72\% & 3.73\% & 41.33\% & 48.93\% & 52.57\% & 45.83\%\tabularnewline
\midrule 
\multirow{3}{*}{StemGNN} & MAE & 19.48 & 21.40 & 24.90 & 21.61 & 19.74 & 22.07 & 26.20 & 22.23 & 14.49 & 15.84 & 18.10 & 15.91 & 1.44 & 1.93 & 2.57 & 1.92 & 21.45 & 23.56 & 24.98 & 22.89\tabularnewline
 & RMSE & 30.74 & 33.46 & 38.29 & 33.80 & 32.32 & 36.16 & 42.32 & 36.46 & 23.02 & 25.38 & 28.77 & 25.44 & 3.12 & 4.38 & 5.88 & 4.46 & 41.09 & 46.95 & 51.97 & 46.21\tabularnewline
 & MAPE & 13.84\% & 15.85\% & 19.50\% & 16.10\% & 8.27\% & 9.20\% & 11.00\% & 9.20\% & 9.73\% & 10.78\% & 12.50 & 10.90\% & 3.08\% & 4.54\% & 6.55\% & 4.54\% & 57.12\% & 65.34\% & 62.81\% & 57.26\%\tabularnewline
\midrule 
\multirow{3}{*}{GMAN} & MAE & 18.27 & \uline{18.81} & \uline{20.01} & \uline{18.83} & 19.25 & 20.33 & \uline{22.25} & 20.43 & 13.80 & \uline{14.62} & \uline{15.72} & 14.81 & 1.34 & 1.65 & \textbf{1.89} & \uline{1.58} & \multirow{1}{*}{} & \multicolumn{2}{c}{N/A} & \tabularnewline
 & RMSE & 29.35 & 30.85 & \textbf{31.32} & 30.93 & 31.20 & \uline{33.30} & \uline{36.40} & \uline{33.30} & 22.88 & 24.12 & 26.47 & 24.19 & 2.92 & 3.81 & \textbf{4.38} & 3.75 & \multirow{1}{*}{} & \multicolumn{2}{c}{N/A} & \tabularnewline
 & MAPE & 12.66\% & 13.25\% & \uline{13.40\%} & 13.21\% & 8.21\% & 8.63\% & \uline{9.48\%} & 8.69\% & 9.41\% & 9.57\% & \uline{10.56\%} & 9.69\% & 2.88\% & 3.71\% & \uline{4.51\%} & 3.69\% & \multirow{1}{*}{} & \multicolumn{2}{c}{N/A} & \tabularnewline
\midrule 
\multirow{3}{*}{MTGNN} & MAE & 18.65 & 19.48 & 20.96 & 19.50 & 19.23 & 20.83 & 23.60 & 20.94 & 14.30 & 15.25 & 16.80 & 15.31 & 1.34 & 1.67 & 1.97 & 1.60 & \uline{16.78} & \uline{18.43} & \uline{20.49} & \uline{18.18}\tabularnewline
 & RMSE & 30.13 & 32.02 & 34.66 & 32.00 & 31.15 & 33.93 & 38.10 & 34.03 & 22.55 & 24.41 & 26.96 & 24.42 & 2.84 & 3.79 & 4.55 & 3.70 & \uline{36.91} & \uline{42.62} & \uline{48.33} & \uline{42.04}\tabularnewline
 & MAPE & 13.32\% & 14.08\% & 14.96\% & 14.04\% & 8.55\% & 9.30\% & 10.10\% & 9.10\% & 10.56\% & 10.54\% & 10.90\% & 10.70\% & 2.80\% & 3.74\% & 4.57\% & 3.57\% & 48.16\% & 51.31\% & 56.25\% & 50.77\%\tabularnewline
\midrule 
\multirow{3}{*}{STNorm} & MAE & 18.28 & 18.92 & 20.20 & 18.96 & 19.15 & 20.63 & 22.60 & 20.52 & 14.44 & 15.53 & 17.20 & 15.54 & 1.34 & 1.67 & 1.96 & 1.60 & 18.74 & 21.14 & 24.05 & 20.69\tabularnewline
 & RMSE & 29.70 & 31.12 & 32.91 & 30.98 & 31.70 & 35.10 & 38.65 & 34.85 & 22.68 & 25.07 & 27.86 & 25.01 & 2.88 & 3.83 & 4.52 & 3.71 & 40.86 & 48.24 & 55.27 & 47.55\tabularnewline
 & MAPE & \uline{12.28\%} & \uline{12.71\%} & 13.43 & \uline{12.69\%} & 8.26\% & 8.84\% & 9.60\% & 8.77\% & 9.32\% & 9.98\% & 11.30\% & 10.03\% & 2.82\% & 3.75\% & 4.62\% & 3.60\% & \uline{32.66\%} & \textbf{37.07\%} & 42.63\% & 35.98\%\tabularnewline
\midrule
\midrule 
\multirow{3}{*}{STID} & MAE & \textbf{17.51} & \textbf{18.29} & \textbf{19.58} & \textbf{18.29} & \textbf{18.31} & \textbf{19.59} & \textbf{21.52} & \textbf{19.54} & \textbf{13.28} & \textbf{14.21} & \textbf{15.58} & \textbf{14.20} & \textbf{1.30} & \textbf{1.62} & \textbf{1.89} & \textbf{1.55} & \textbf{16.08} & \textbf{17.87} & \textbf{19.25} & \textbf{17.39}\tabularnewline
 & RMSE & \textbf{28.48} & \textbf{29.86} & \uline{31.79} & \textbf{29.82} & \textbf{30.39} & \textbf{32.90} & \textbf{36.29} & \textbf{32.85} & \textbf{21.66} & \uline{23.57} & \textbf{25.89} & \textbf{23.49} & \uline{2.81} & \textbf{3.72} & \uline{4.40} & \textbf{3.62} & \textbf{34.49} & \textbf{41.65} & \textbf{45.77} & \textbf{40.80}\tabularnewline
 & MAPE & \textbf{12.00\%} & \textbf{12.46\%} & \textbf{13.38\%} & \textbf{12.49\%} & \textbf{7.72\%} & \textbf{8.30\%} & \textbf{9.15\%} & \textbf{8.25\%} & \textbf{8.62\%} & \textbf{9.24\%} & \textbf{10.33\%} & \textbf{9.28\%} & \uline{2.73\%} & \uline{3.68\%} & \textbf{4.47\%} & \textbf{3.51\%} & \textbf{31.95\%} & 37.80\% & \uline{40.26\%} & \textbf{35.53\%}\tabularnewline
 \midrule
\bottomrule
\end{tabular}
}
\end{table*}

\begin{figure*}
    \setlength{\abovecaptionskip}{0.1cm}
    \setlength{\belowcaptionskip}{-0.2cm}
  \centering
  \includegraphics[width=0.965\linewidth]{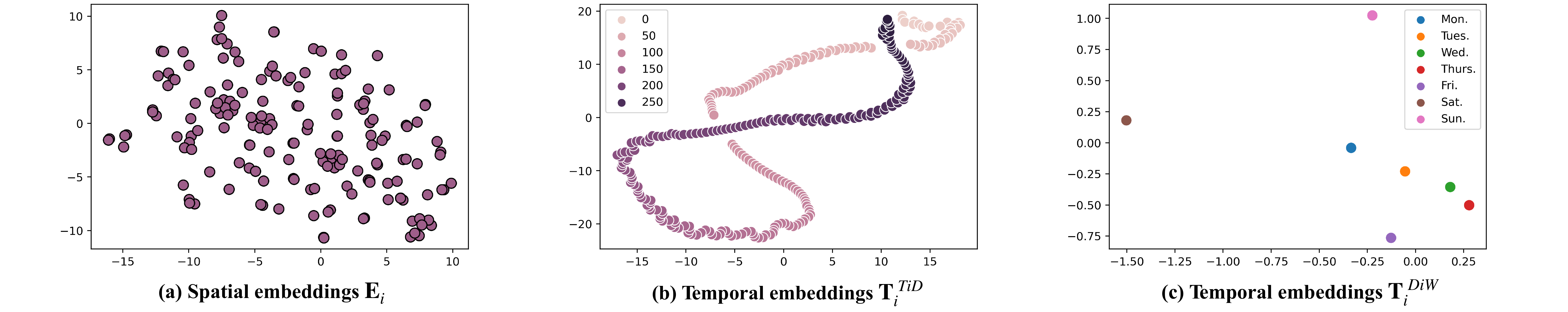}
  \caption{Visualization of learned spatial and temporal identities.}
  \label{vis}
\end{figure*}

\subsection{Visualization}
In order to further intuitively understand and evaluate our model, in this section, we visualize the learned spatial and temporal identities.
Specifically, we visualize $\mathbf{E}\in\mathbb{R}^{N\times D}$, $\mathbf{T}^{\text{TiD}}\in\mathbb{R}^{N_d\times D}$, and $\mathbf{T}^{\text{DiW}}\in\mathbb{R}^{N_w\times D}$ of STID on PEMS08 datasets, where $N=170$, $N_d=288$, and $N_w=7$.
Here we utilize t-SNE~\cite{2008t-SNE} to visualize $\mathbf{E}$ and $\mathbf{T}^{\text{TiD}}$.
For $\mathbf{T}^{\text{DiW}}\in\mathbb{R}^{N_w\times D}$, where $N_w=7 \ll D=32$, we train STID by setting the embedding size of $\mathbf{T}^{\text{DiW}}$ to 2 to get a more accurate visualization.
The results are shown in Figure \ref{vis}.

First, Figure \ref{vis}(a) demonstrates that the identities of different variables~(\ie time series) are likely to cluster.
This is in line with the characteristics of the transportation system.
For example, nearby traffic sensors in the road network tend to share similar patterns.
Second, Figure \ref{vis}(b) visualize the embeddings of 288 time slots for each day.
It is obvious that there is daily periodicity in the PEMS08 dataset.
Moreover, adjacent time slots tend to share similar identities.
%
Last, Figure \ref{vis}(c) shows that the identities of weekdays are similar, while the weekends' are very different.

\section{Conclusion}
In this paper, we propose to explore the critical factors of MTS forecasting to design a model that is as powerful as STGNNs but more concise and effective.
Specifically, we identify the indistinguishability of samples in both spatial and temporal dimensions as a key bottleneck.
Subsequently, we propose a simple yet effective baseline for MTS forecasting by attaching spatial and temporal identity information, \ie STID.
STID achieves better efficiency and performance simultaneously based on simple networks.
These results suggest that by solving the indistinguishability of samples, we can design models more freely, without being limited to STGNNs.



\begin{acks}
This work was supported in part by the National Natural Science Foundation of China under Grant No. 61902376, No. 61902382, and No. 61602197, in part by CCF-AFSG Research Fund under Grant No. RF20210005, and in part by the fund of Joint Laboratory of HUST and Pingan Property \& Casualty Research (HPL).
In addition, Zhao Zhang is supported by the China Postdoctoral Science Foundation under Grant No. 2021M703273.
\end{acks}

\bibliographystyle{ACM-Reference-Format}
\normalem
\bibliography{references}

\end{document}